\documentclass[10pt, conference, compsocconf]{IEEEtran}

\usepackage{times}
\usepackage{epsfig}
\usepackage{graphicx}
\usepackage{amssymb}
\usepackage{epstopdf}
\usepackage{array}

\usepackage{listings}
\usepackage{xcolor,pifont}
\newcommand*\colourcheck[1]{%
  \expandafter\newcommand\csname #1check\endcsname{\textcolor{#1}{\ding{52}}}%
}
\colourcheck{green}

\usepackage{colortbl}

\usepackage{multirow}
\usepackage{algorithm}
\usepackage{algorithmic}
\usepackage{wrapfig, blindtext}
\usepackage{amsmath}

\usepackage{url}

\begin{document}
\title{Understanding the impact of image and input resolution\\on deep digital pathology patch classifiers}


\author{\IEEEauthorblockN{Eu Wern Teh}
\IEEEauthorblockA{School of Engineering, University of Guelph\\
Vector Institute for Artificial Intelligence\\
eteh@uoguelph.ca}
\and
\IEEEauthorblockN{Graham W. Taylor}
\IEEEauthorblockA{School of Engineering, University of Guelph\\
Vector Institute for Artificial Intelligence\\
gwtaylor@uoguelph.ca}
}

\maketitle

\begin{abstract}
We consider annotation efficient learning in Digital Pathology (DP), where expert annotations are expensive and thus scarce. We explore the impact of image and input resolution on DP patch classification performance. We use two cancer patch classification datasets PCam and CRC, to validate the results of our study. Our experiments show that patch classification performance can be improved by manipulating both the image and input resolution in annotation-scarce and annotation-rich environments.
 We show a positive correlation between the image and input resolution and the patch classification accuracy on both datasets.
By exploiting the image and input resolution, our final model trained on $\textless$ 1\% of data performs equally well compared to the model trained on 100\% of data in the original image resolution on the PCam dataset.

\end{abstract}


\IEEEpeerreviewmaketitle

\section{Introduction}

Digital Pathology (DP) is a medical imaging field where microscopic images are analyzed to perform Digital Pathology tasks (e.g., cancer diagnosis). The appearance of hardware for a complete DP system popularized the usage of whole slide images (WSI) \cite{pantanowitz2010digital}. These whole slide images are high-resolution images, usually gigapixel in size. Without resizing WSIs, these high-resolution images will not fit in the memory of off-the-shelf GPUs. On the other hand, resizing WSIs destroys fine-grained information crucial in DP tasks, such as cancer classification. A common practice is to divide WSIs into small patches to address this issue.
Most of the DP solutions for patch classification use single-scale images, where a fixed zoom-level of WSI is used~\cite{veeling2018rotation,kather2019predicting,aresta2019bach,srinidhi2020deep}. Therefore, a standard augmentation strategy such as Random-resized-crop will not work in patch classification  (more details are provided in Section~\ref{sec:rrc}).

\begin{figure}[htb]
  \centering
  \includegraphics[width=3.3in,   ]{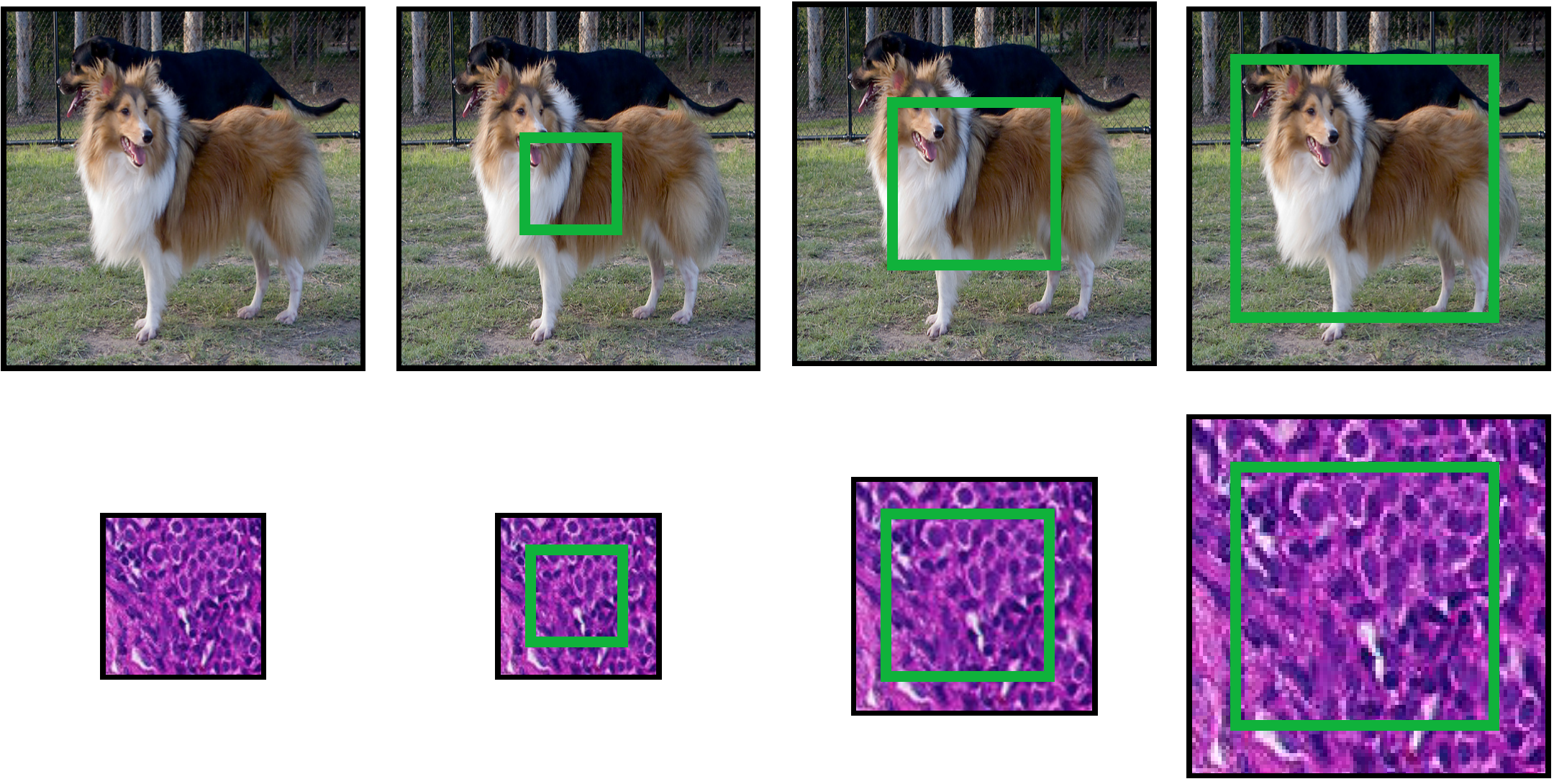}
  \caption{
  An illustration of the differences between input and image resolutions. In both rows of images, input resolution increases from left to right. However, only the bottom row has an increase in image resolution. In Digital Pathology, an increase in both the image and input resolution allows a model to capture fine-grained information without losing the global context of a given image. 
  }
  \label{fig:main_diff}
\end{figure}

\begin{figure}[!htb]
  \centering
  \includegraphics[width=3.3in,   ]{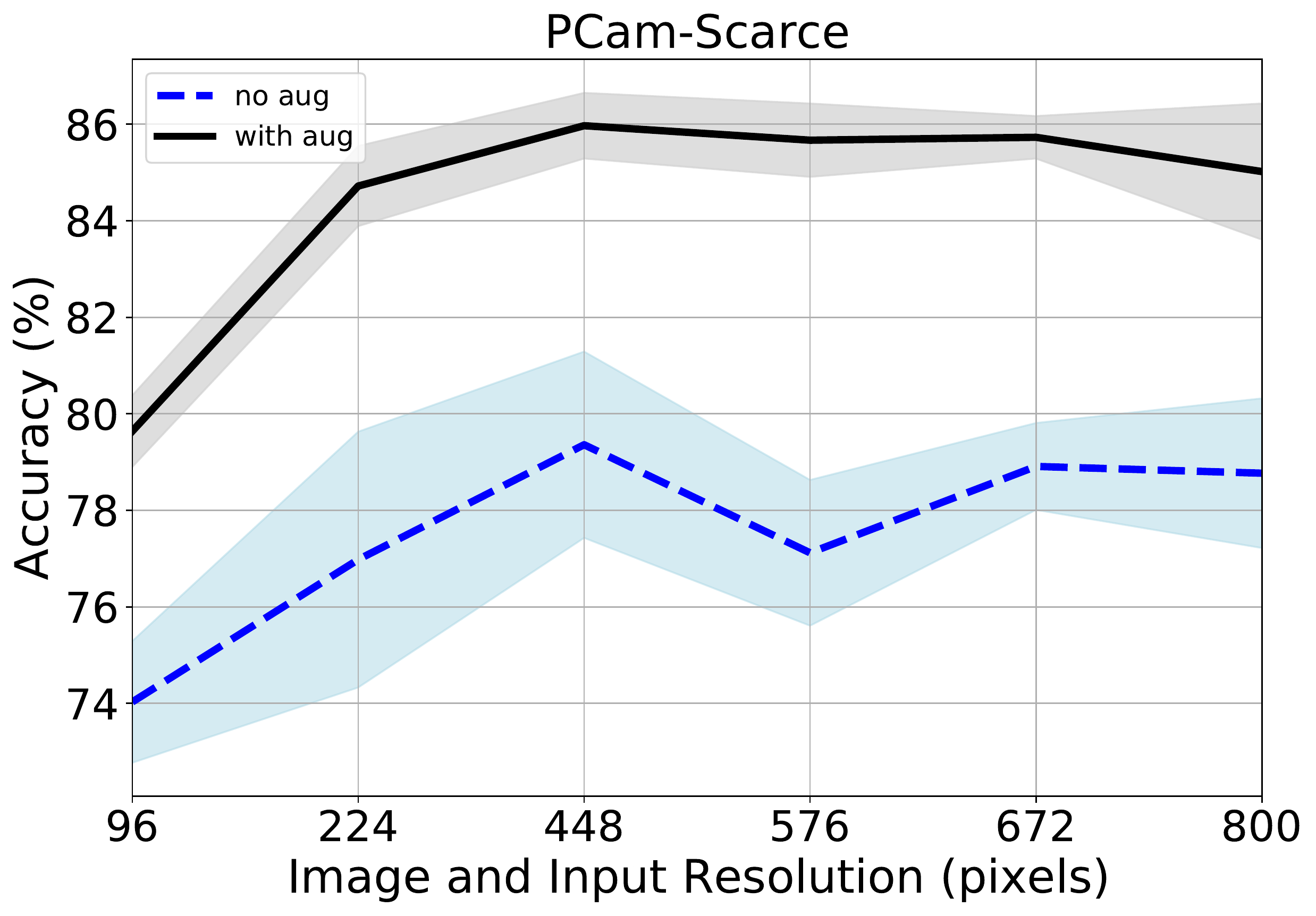}
  \caption{
Image and input resolution effects on models trained with $\textless$ 1\% of the original dataset.
  The shaded areas represent one standard deviation of uncertainty. The dotted line represents models trained without data augmentation. 
  }
  \label{fig:pcam_1p}
\end{figure}

The Deep Learning field has focused on innovating architectures ranging from AlexNet to ResNet to Transformers \cite{krizhevsky2012imagenet,he2016deep,dosovitskiy2020image}. ``The bigger the model, the better" approach works when a massive quantity of human annotations is available. However, in an annotation-scarce environment, models with higher capacity often perform worse due to overfitting~\cite{goodfellow2016deep}.
We take a step back from designing complex methods and explore image and input resolution factors, which are often ignored as part of an input processing pipeline.
We show that we can improve patch classification with a fixed zoom level by increasing the image and input resolution of image patches in DP (Figure~\ref{fig:pcam_1p}). By increasing the image and input resolution, we allow our model to focus on fine-grained information without losing coarse-grained information, resulting in an increase of 6.33\% (PCam-Scarce) and 9.83\% (CRC-Scarce) in patch classification accuracy.

\section{Related work}\label{sec:rel}

\noindent{\textbf{Annotation-efficient learning:}} In the Digital Pathology (DP) domain, the availability of human labels is often scarce due to the high annotation costs \cite{srinidhi2020deep}. This data scarcity is a challenge for supervised deep learning models as these models often require tremendous amounts of human labels to be effective \cite{goodfellow2016deep}. An effective way of combating label scarcity is via transfer learning in the form of pre-training \cite{huh2016makes}. Transfer learning from a pre-trained ImageNet model is shown to be effective in the DP domain, especially in the low-data regime~\cite{hegde2019similar,teh2020learning,kupferschmidt2021strength}.
\\

\noindent{\textbf{The effect of input resolution on model performance:}} There is an important distinction between image resolution and input resolution (Figure~\ref{fig:main_diff}). Image resolution refers to the width and height of an image in pixels, while input resolution refers to the input width and height being fed to a model. Random-resized-crop augmentation is a standard input augmentation technique used in the Natural Image (NI) domain for image classification~\cite{touvron2019fixing,tan2019efficient,he2016deep}. During Random-resized-crop augmentation, the image resolution remains the same throughout training, while the crop area changes~\cite{rrc}. The crop area is randomly scaled between 0.08 to 1.00 of the image resolution, followed by another random re-scaling between 0.75 to 1.33. Finally, the crop area is resized to the given input resolution.

In the NI domain,  classification performance can be improved by increasing the input resolution (crop size)~\cite{touvron2019fixing,tan2019efficient}.
Touvron et al. discover that different training and testing input resolution affects model performance~\cite{touvron2019fixing}. The most optimal train and test input resolution for the ImageNet dataset are 384$\times$384 and 448$\times$448.  Tang et al. push the limits of their EfficientNet model by increasing the input resolution of the images~\cite{tan2019efficient}. Teh et al. show that input resolution has a large influence on image retrieval performance~\cite{teh2020proxynca++}.
It is reasonable to use Random-resized-crop in the NI domain, as NI images come with a wide range of different image resolutions (Figure~\ref{fig:imagenet}). However, in the DP domain, Random-resized-crop is ineffective for patch classification because image patches have the same image resolution (fixed zoom-level)  throughout the dataset (Section~\ref{sec:rrc}).

\section{Methodology and Control Factors}

\noindent{\textbf{Supervised-training:}} A ResNet-34 architecture is used as our model in all of our experiments\footnote{With the exception of experiments in Table~\ref{table:exp_colon_pcam_small} where we explore ResNet-152 and ResNeXt-101 architectures \cite{xie2017aggregated}.} where we trained them with cross-entropy loss as described in Equation~\ref{eq:1} \cite{he2016deep}. $B$ denotes the batch size, $x_i$ denotes a single image with a corresponding image label, $y_i$, and $K$ denotes the number of classes in the respective dataset. $f(x_i)$ represents a deep model that accepts an image, $x_i$ and returns a vector of size $K$ denoting the classification scores. 

\begin{align}
L  =  -\frac{1}{B}\sum_{i}^{B}\log\frac{\text{exp}(f(x_{i})[y_{i}])}{\sum_{k=1}^{K}\text{exp}(f(x_{i})[k])}\label{eq:1}
\end{align} \\

\noindent{\textbf{Model initialization:}} In Section~\ref{sec:scarce} and \ref{sec:rich}, we use randomly initialized weights in our experiments. In Section~\ref{sec:limit}, we use both randomly initialized and ImageNet pre-trained ResNet-34 model to explore the limits of annotation-efficient learning for patch classification in DP. \\

\begin{figure*}[htb]
  \centering
  \includegraphics[width=6.40in,   ]{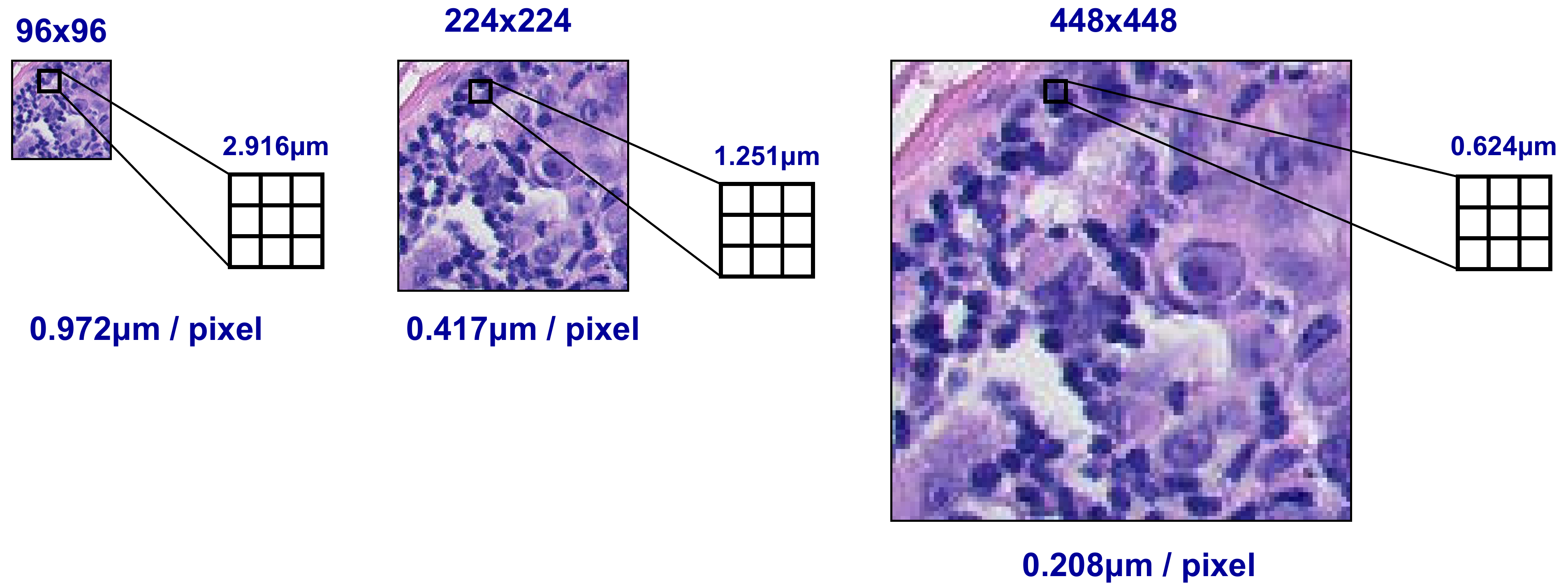}
  \caption{
  An illustration of image resolution and the corresponding receptive fields of a 3$\times$3 convolution layer at various image resolutions. The leftmost image is resized twice (224$\times$224  and 448$\times$448) via bilinear interpolation from an original image at 96$\times$96 resolution with $0.972\mu m$/pixel. As image resolution increases, the receptive field decreases, allowing finer-grained features to be captured by the convolution layer.
  }
  \label{fig:main}
\end{figure*}

\noindent{\textbf{Data augmentation:}} We perform data augmentation via random rotation, random cropping, random horizontal flipping, and color jittering following Teh et al.~\cite{teh2020learning} (Algorithm~\ref{alg:code}). Each image is resized with matching height, $h$ and width, $w$ via bilinear interpolation. As image resolution increases, the receptive field of each convolution layer in the model increases, resulting in model awareness towards finer-grained details in the images (Figure~\ref{fig:main}). 
%Table~\ref{table:image_res} shows the image resolution and corresponding pad size used in our experiments. The pad size is approximately 12.5\% to 14.3\% of the corresponding image resolution. 
Additionally, we also perform experiments without data augmentation in Section~\ref{sec:scarce} (Algorithm~\ref{alg:code_no}). \\

\noindent{\textbf{Dataset demography:}} We use the Patch Camelyon (PCam) and the Colorectal cancer (CRC) dataset in our experiments. For each dataset, we create two subsets: scarce and full. The scarce subset consists of $\textless 1$ \% of the dataset, while the full subset consists of $100\%$ of the dataset.

The PCam dataset consists of 262,144 training images, 32,768 validation images, and 32,768 test images~\cite{veeling2018rotation}. It comprises two classes: normal tissue and tumor tissue. Each image has a dimension of 96$\times$96 pixels with a resolution of 0.972$\mu$m per pixel. On the scarce subset of PCam, we randomly extract 2000 images from the training set and 2000 images from the validation set. The 2000 validation images are used in both scarce and full experiments to determine early stopping. There is approximately 0.76\% of images in PCam-Scarce training set when compared to the full training set.

The CRC dataset consists of 100,000 images, and we split the dataset into 70\% training set, 15\% validation set, and 15\% test set following Kather et al.~\cite{kather2019predicting}. It consists of nine classes: adipose tissue, background, debris, lymphocytes, mucus, smooth muscle, normal colon tissue, cancer-associated stroma, and colorectal adenocarcinoma epithelium. Each image has a dimension of 224$\times$224 pixels with a resolution of 0.5$\mu$m per pixel. On the scarce subset of CRC, we randomly extract 693 images (77 images per class) from the training set and 693 images (77 images per class) from the validation set. The 693 validation images are used in both scarce and full experiments to find the best epoch at which to early stop. There is approximately 0.99\% of images in CRC-Scarce training set when compared to the full training set. \\

\noindent{\textbf{Additional Experimental Settings:}} We train all our models on the scarce subset for 100 epochs and 20 epochs for the full subset of the corresponding dataset. 
We use a validation set to select the best epoch and a separate test set is used for model evaluation.
All models are trained with the Adam optimizer with a learning rate of $1e^{-4}$, a weight decay of $5e^{-3}$, and a batch size of 32. We train all models five times with different seeds~\footnote{Random seeds affect the weight initialization of our models.} and report the mean accuracy with one standard deviation of uncertainty. Furthermore, we also use the SciPy Pearson library to compute the Pearson Correlation coefficient between test accuracy and the input and image resolution.  

\section{Experiments}

We examine the effects of image and input resolution in three different settings: Annotation-scarce environments (\S~\ref{sec:scarce}), Annotation-rich environments (\S~\ref{sec:rich}), and Transfer Learning settings (\S~\ref{sec:limit}). \\ \\ \\

\subsection{The effects of image and input resolution on annotation-scarce dataset}~\label{sec:scarce}

%Figure~\ref{fig:pcam_lcrc_1p_noaug} shows our experiments on PCam-Scarce and CRC-Scarce datasets with and without data augmentation. On models with augmentation, there is a strong positive correlation of 0.703 (PCam-Scarce) and 0.944 (CRC-Scarce) between the image-and-input resolution and patch accuracy. Similarly, there is also a strong correlation of 0.771 (PCam-Scarce) and 0.938 (CRC-Scarce) on models without augmentation.

%Models performances suffer in general without data augmentation on the PCam-Scarce dataset. Nevertheless, on the CRC-Scarce dataset, models without data augmentation surpass models with data augmentation as image and input resolution reach 576 and beyond.

\begin{figure}[!htb]
  \centering
  \includegraphics[width=3.3in,   ]{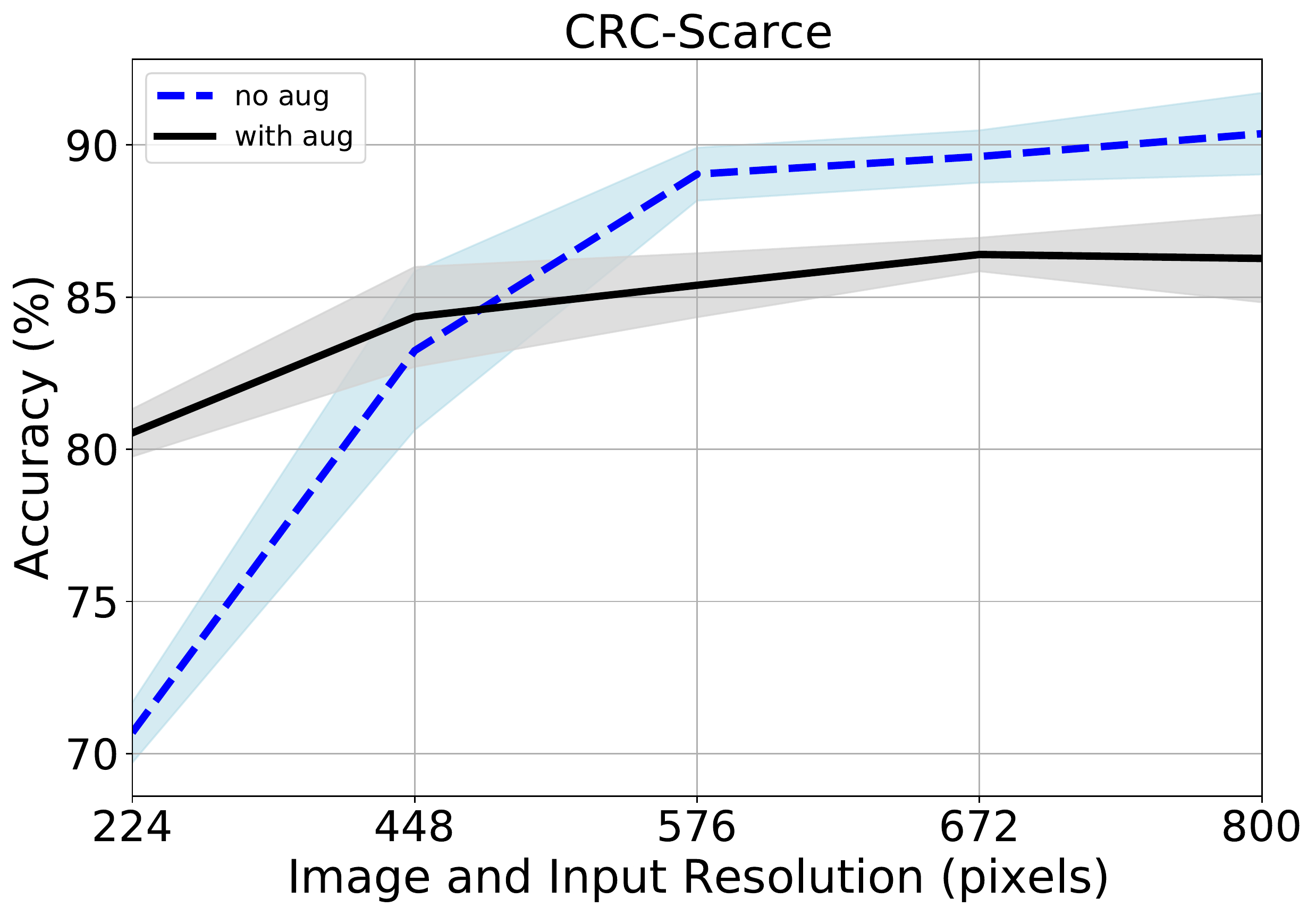}
  \caption{
Image and input resolution effects on models trained with $\textless$ 1\% of the original dataset.
  The shaded areas represent one standard deviation of uncertainty. The dotted line represents models trained without data augmentation. 
  }
  \label{fig:lcrc_1p}
\end{figure}

We analyze the impact of image and input resolution on two annotation-scarce datasets: PCam-Scarce and CRC-Scarce with and without data augmentation (Figures~\ref{fig:pcam_1p} and \ref{fig:lcrc_1p}). On models with augmentation, there is a positive correlation of 0.703 (PCam-Scarce) and 0.944 (CRC-Scarce) between the image and input resolution and patch accuracy. Similarly, there is also a correlation of 0.771 (PCam-Scarce) and 0.938 (CRC-Scarce) on models without augmentation. Models' performance suffers in general without data augmentation on the PCam-Scarce dataset. Nevertheless, on the CRC-Scarce dataset, models without data augmentation surpass models with data augmentation as image and input resolution approach 576$\times$576 and beyond. This result shows that an increase of image and input resolution alone may be sufficient to regularize the network on the CRC-Scarce dataset. Additionally, this result also shows that too much regularization could hurt generalization on test set.

\begin{table*}[h]
\centering
\setlength{\tabcolsep}{23pt}
\scalebox{1.0}{
\begin{tabular}{|lcc|c|c|} \hline
\multicolumn{3}{|c|}{}&\multicolumn{1}{c|}{PCam-Scarce}&\multicolumn{1}{c|}{CRC-Scarce} \\ \hline
Models & Resolution & GFLOPs & Accuracy (\%) & Accuracy (\%)  \\ \hline
ResNet-34 & 96$\times$96 &  0.68 & \cellcolor{gray!24}79.64$\pm$0.74 & 72.17$\pm$1.65  \\
ResNet-34 & 224$\times$224 &  3.68 & 84.72$\pm$0.83 & \cellcolor{gray!24}80.54$\pm$0.78  \\
ResNet-34 & 448$\times$448 &  14.73 & \textbf{85.97$\pm$0.68} & 84.35$\pm$1.64  \\
ResNet-34 & 576$\times$576 &  24.35 & 85.67$\pm$0.76 & 85.39$\pm$1.05  \\
ResNet-34 & 672$\times$672 &  33.14 & 85.73$\pm$0.44 & \textbf{86.40$\pm$0.55}  \\
ResNet-34 & 800$\times$800 &  46.97 & 85.02$\pm$1.41 & 86.27$\pm$1.44 \\
\hline
ResNet-152 & 96$\times$96 &  2.14 & \cellcolor{gray!24}77.47$\pm$1.57 & 55.68$\pm$1.00  \\
ResNet-152 & 224$\times$224 &  11.63 & 79.97$\pm$1.10 & \cellcolor{gray!24}65.34$\pm$1.44 \\
\hline
ResNeXt-101 & 96$\times$96 &  3.04 & \cellcolor{gray!24}79.34$\pm$0.60 & 61.53$\pm$0.89 \\
ResNeXt-101 & 224$\times$224 &  16.55 & 80.06$\pm$0.77 & \cellcolor{gray!24}71.21$\pm$1.67 \\
\hline
\end{tabular}
}
%\bigskip
    \caption{Patch classification accuracy of our models trained on the PCam-Scarce dataset and CRC-Scarce dataset. All models are trained with data augmentation. Column 2 denotes both the image and input resolution. The shaded cells represent the default configurations on both datasets.}
    \label{table:exp_colon_pcam_small}
\end{table*}

Table~\ref{table:exp_colon_pcam_small} shows our experimental results on the PCam-Scarce and CRC-Scarce dataset at various image and input resolutions. In addition to the ResNet-34 architecture, we also experiment with ResNet-152 and ResNeXt-101 (32\texttimes 8d) architectures. ResNet-152 is a direct upgrade of ResNet-34, with more convolution layers. ResNeXt is an enhanced version of the corresponding ResNet architecture with increased parallel residual blocks and more channels  \cite{xie2017aggregated}. ResNeXt-101 (32\texttimes 8d) has 32 parallel residual blocks with 8\texttimes\ more channels compared to the vanilla ResNet-101 architecture. Additionally, we also report the number of floating-point operations (GFLOPS) for each model at various image and input resolutions\footnote{We exclude the final linear classifier in this calculation as it is held constant across architectures.}.

\begin{figure*}[!htb]
  \centering
  \includegraphics[width=6.40in,   ]{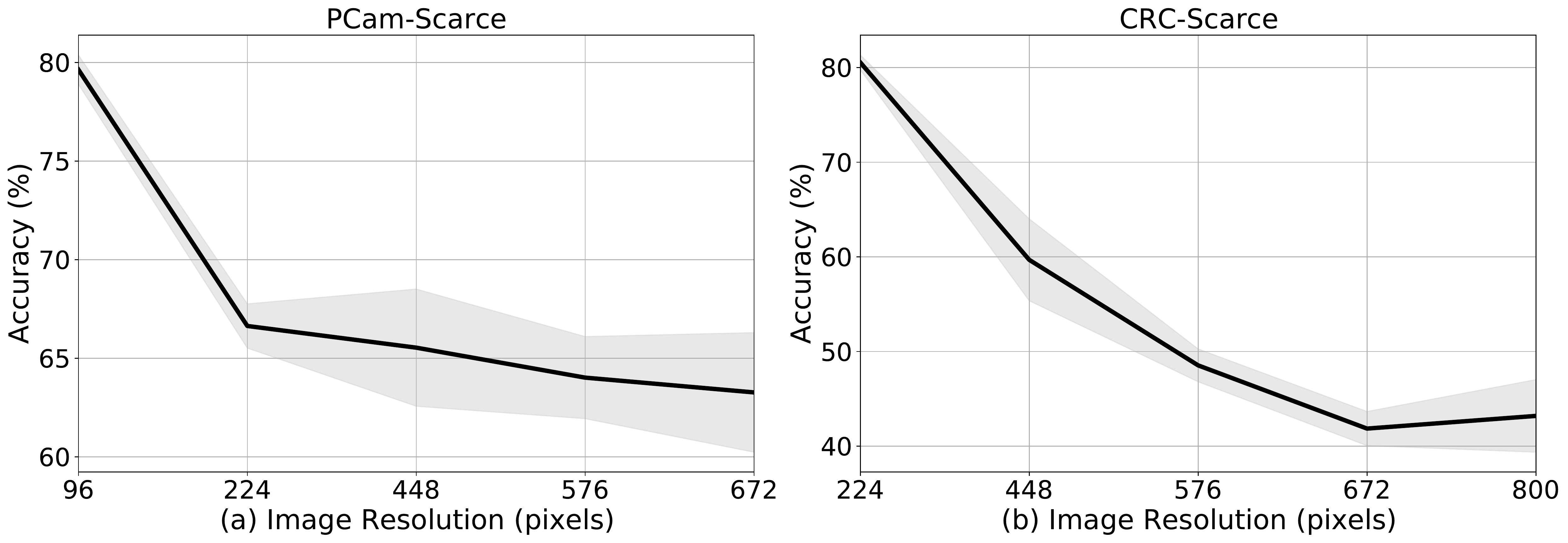}
  \caption{Effect of image resolution on models trained with $\textless$ 1\% of the original dataset. Input resolution is held constant in this experiment. The shaded areas represent one standard deviation of uncertainty. All models train with data augmentation.
  %Randomly initialized weights are set to all ResNet-34 models in this experiment.
  }
  \label{fig:pcam_lcrc_1p_alt}
\end{figure*}

 Model performance decreases when larger and more complex architectures (ResNet-152 and ResNeXt-101) are used on the annotation-scarce dataset (Table~\ref{table:exp_colon_pcam_small}). This result concurs with Goodfellow et al., where higher capacity models can overfit the dataset~\cite{goodfellow2016deep}. On the other hand, increasing the image and input resolution yields a performance gain, while models with higher capacity failed to do so. This result suggests that increasing the image and input resolution may indeed be regularizing high capacity models in the low-annotation setting.
 However, increasing the image and input resolution also leads to increased computational cost, which grows quadratically with those factors.

Figure~\ref{fig:pcam_lcrc_1p_alt} shows that varying the image resolution alone is not sufficient. There is a negative correlation of 0.835 (Pcam-Scarce) and 0.970 (CRC-Scarce) between image resolution and patch classification accuracy. As the image resolution increases, a model can capture finer-grained information. Nonetheless, if we do not increase the input resolution proportionally, the model loses global information, causing a significant drop in performance. \\

\subsection{The effects of image and input resolution on an annotation-rich dataset}~\label{sec:rich}

\begin{figure*}[htb]
  \centering
  \includegraphics[width=6.40in,   ]{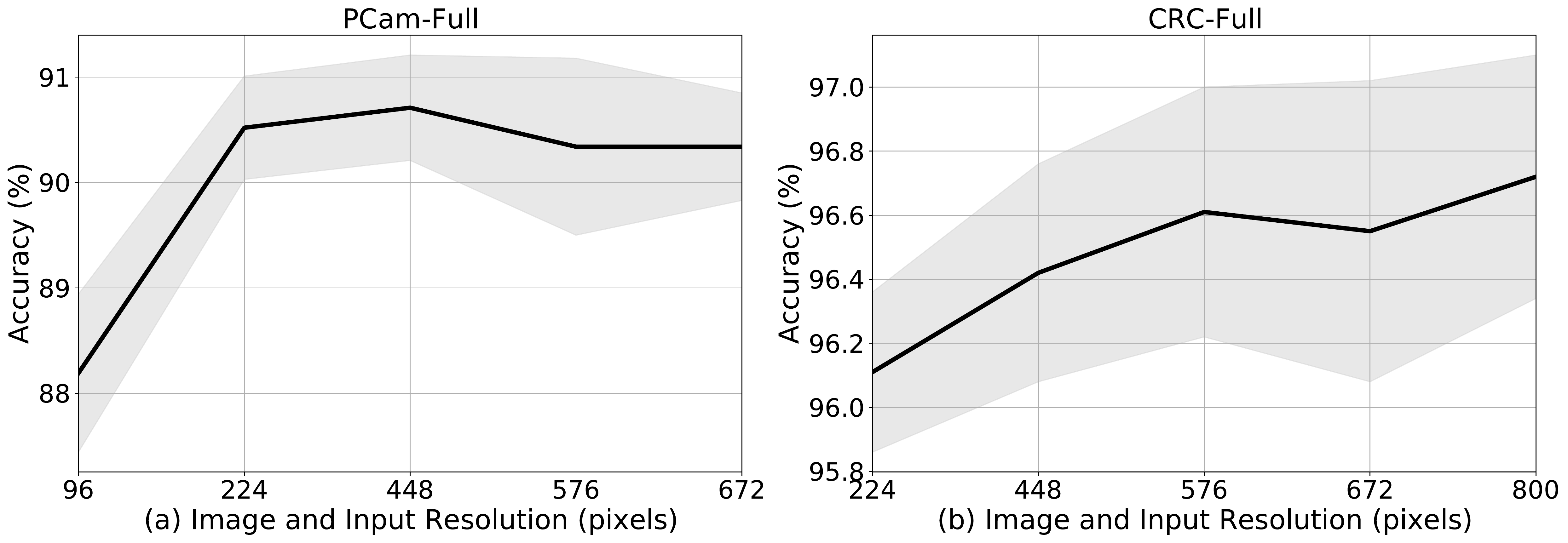}
  \caption{Image and input resolution effects on models trained with 100\% of the original dataset. The shaded areas represent one standard deviation of uncertainty. All models are trained with data augmentation.
  %Randomly initialized weights are set to all ResNet-34 models in this experiment.
  }
  \label{fig:pcam_lcrc_full}
\end{figure*}

We examine the impact of image and input resolution on two annotation-rich datasets: PCam-Full and CRC-Full.
Figure~\ref{fig:pcam_lcrc_full} shows that as we increase the image and input resolution, the models' performance generally increases.
There is a positive correlation of 0.649 (PCam-Full) and 0.961 (CRC-Full) between the image and input resolution and patch classification accuracy.
For the PCam-Full dataset, peak performance occurs at 576$\times$576 resolution. The model achieves peak performance at 800$\times$800 resolution for the CRC-Full dataset. There is a gain of 2.52\% in accuracy for the PCam-Full dataset and a gain of 0.62\% in accuracy for the CRC-Full dataset when compared to the accuracy of models trained with the original image resolution of the respective datasets (PCam-Full: 96$\times$96, CRC-Full: 224$\times$224). 
%These results show that both the image and input resolution have a smaller impact on the model's performance in an annotation-rich environment.

\subsection{Annotation-efficient learning in the transfer learning setting}~\label{sec:limit}

\begin{figure*}[htb]
  \centering
  \includegraphics[width=6.40in,   ]{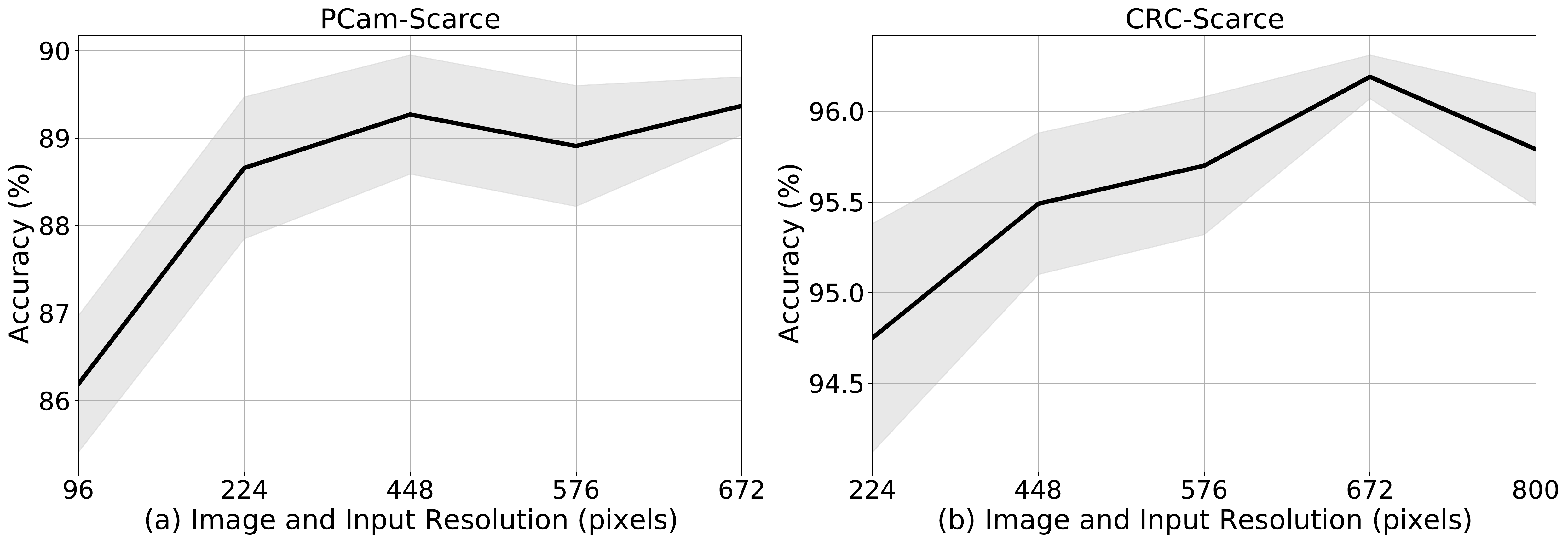}
  \caption{Image and input resolution effects on models trained with $\textless$ 1\% of the original dataset.
  The shaded areas represent one standard deviation of uncertainty. PCam-Scarce models are trained with data augmentation, but CRC-Scarce models are trained without data augmentation. All models are initialized with ImageNet pre-trained weights.
  %Randomly initialized weights are set to all ResNet-34 models in this experiment.
  }
  \label{fig:pcam_lcrc_1p_imgnet}
\end{figure*}

Transfer learning in the form of pre-training is a common strategy to tackle annotation efficient learning (\S~\ref{sec:rel}). In the previous experiments we avoided transfer learning to isolate the contribution of image and input resolution on performance. In this experiment, we study the interaction between image and input resolutions and pre-trained features (Table~\ref{table:exp_colon_pcam} and Figure~\ref{fig:pcam_lcrc_1p_imgnet}). Previously, we observed that on the PCam-Scarce and CRC-Scarce datasets, there is a gain of up to 6.33\% and 9.83\% in patch classification accuracy by increasing the image and input resolution.  Here, we observe an additional gain of 3.30\% and 5.42\% in patch classification accuracy when we initialize the weights of the ResNet-34 model by supervised pre-training on the ImageNet 2012 dataset. On the PCam-Scarce dataset, our final model performs equally well compared to the model trained on 100\% of the dataset in the original image resolution (96$\times$96). For the CRC-Scarce dataset, our final model is only 2.22\% away from the model trained on 100\% of the dataset in the original image resolution (224$\times$224). 
Additionally, there is a positive correlation of 0.810 (PCam-Scarce) and 0.875 (CRC-Scarce) between the image and input resolution and the patch classification accuracy on models initialized with ImageNet pre-trained weights (Figure~\ref{fig:pcam_lcrc_1p_imgnet}).
This result shows that pre-trained models work well with the image and input resolution factors, yielding an additive gain in performance.

\begin{table*}[!htb]
\centering
\setlength{\tabcolsep}{6pt}
\scalebox{0.9}{
\begin{tabular}{|ccccc|ccccc|} \hline
\multicolumn{5}{|c|}{PCam dataset}&\multicolumn{5}{c|}{CRC dataset} \\ \hline
Subset & Initialization & Augumentation & Resolution & Accuracy (\%) &   Subset & Initialization & Augumentation & Resolution & Accuracy (\%)  \\ \hline
 Scarce &  Random & \greencheck & 96$\times$96 & 79.64$\pm$0.74 &  Scarce &  Random & \greencheck & 224$\times$224 & 80.54$\pm$0.78 \\
 Scarce &  Random & \greencheck & 448$\times$448 & 85.97$\pm$0.68 &  Scarce &  Random & \textcolor{red}{\ding{53}} & 800$\times$800 & 90.37$\pm$1.34 \\
 Scarce &  ImageNet & \greencheck & 448$\times$448 & \textbf{89.27$\pm$0.68} &  Scarce &  ImageNet & \textcolor{red}{\ding{53}} & 800$\times$800 & 95.79$\pm$0.31 \\
%Self-Training & 0.76 &  ImageNet & 448 & - & Self-Training & 1.00 &  ImageNet & 672 & -\\
%\\
\hline
 Full &  ImageNet & \greencheck & 96$\times$96 & 88.88$\pm$0.92 &  Full &  ImageNet & \greencheck & 224$\times$224 & \textbf{98.01$\pm$0.14} \\
\hline
\end{tabular}
}
%\bigskip
    \caption{Accuracy of our models trained on the PCam dataset and CRC dataset. We also show dataset subset (column 1 and 6), model initialization (columns 2 and 7), the use of data augmentation during training (columns 3 and 8) as well as the image and input resolution (columns 4 and 9) in this table.
    }
    \label{table:exp_colon_pcam}
\end{table*}

%\section{Additional Experimental Details}
\subsection{ImageNet 2012 statistics}~\label{sec:imagenetstats}

\begin{figure*}[htb]
  \centering
  \includegraphics[width=6.40in,   ]{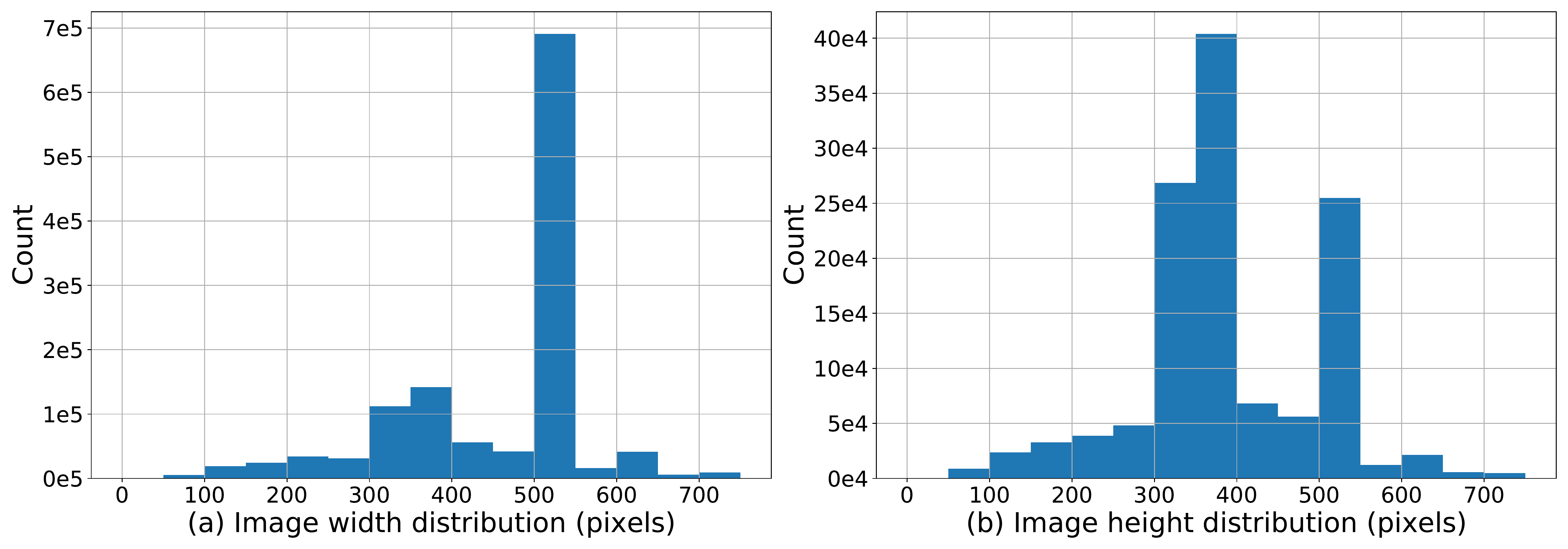}
  \caption{Image resolution distribution of ImageNet 2012 dataset. The average image width and height are 472 and 405, with a standard deviation of 208 and 179.
  }
  \label{fig:imagenet}
\end{figure*}

Figure~\ref{fig:imagenet} shows the image resolution distribution of the ImageNet 2012 dataset. The average image width and height are 472 and 405, with a standard deviation of 208 and 179. $54\%$ of the images have an image width of 500 to 550, and $52\%$ have an image height of 300 to 400. The large variation in image resolution makes Random-Resized Crop a suitable image augmentation strategy for the ImageNet 2012 dataset. The aim of the Random-Resized Crop algorithm is to generalize a model to unseen images of different image resolutions. 

\subsection{Data augmentation for Digital Pathology}~\label{sec:rrc}

Algorithm~\ref{alg:code} shows the data augmentation strategy specifically designed for patch classification in Digital Pathology, following Teh et al.~\cite{teh2020learning}. Table~\ref{table:image_res} shows the image and input resolution and the corresponding pad size used in our experiments.
The pad size is approximately 12.5\% to 14.3\% of the corresponding image resolution.

\begin{algorithm}[htb]
\caption{Data Augmentation, PyTorch-like}
\label{alg:code}
\definecolor{codeblue}{rgb}{0.25,0.5,0.5}
\definecolor{codekw}{rgb}{0.85, 0.18, 0.50}
\lstset{
  backgroundcolor=\color{white},
  basicstyle=\fontsize{7.5pt}{7.5pt}\ttfamily\selectfont,
  columns=fullflexible,
  breaklines=true,
  captionpos=b,
  commentstyle=\fontsize{7.5pt}{7.5pt}\color{codeblue},
  keywordstyle=\fontsize{7.5pt}{7.5pt}\color{codekw},
}
\begin{lstlisting}[language=python]
# h,w: height, width; p: padding size
import torchvision.transforms as t

transform = {'train':t.Compose([
                        t.Resize((h, w)),
                        t.Pad(p, padding_mode='reflect'),
                        t.RandomRotation([0, 360]),
                        t.RandomCrop((h,w)),
                        t.RandomHorizontalFlip(0.5),
                        t.ColorJitter(
                            hue= 0.4,
                            saturation=0.4,
                            brightness=0.4,
                            contrast=0.4),
                        t.ToTensor(),
                        ]),
                 'test':t.Compose([
                        t.Resize((h, w)),
                        t.ToTensor(),
                        ])}
\end{lstlisting}
\end{algorithm}

\begin{table}[h]
\centering
\setlength{\tabcolsep}{5pt}
\scalebox{1.0}{
\begin{tabular}{|l|c|c|c|c|c|c|} \hline
Image Resolution (pixels)   & 96 & 224 & 448 & 576 & 672 & 800 \\
\hline
Image Padding (pixels) & 12 & 32 & 64 & 80 & 96 & 114 \\
\hline
\end{tabular}
}
%\bigskip
    \caption{Image resolution and the corresponding pad size.}
    \label{table:image_res}
\end{table}

%\subsection{Random-resized-crop}~\label{sec:rrc}

Algorithm~\ref{alg:code_rrc} shows a typical data augmentation strategy used in the Natural Image domain. We compare our data augmentation strategy (Algorithm~\ref{alg:code}) with respect to Algorithm~\ref{alg:code_rrc} on PCam-Scarce dataset with the image resolution of 96$\times$96 and a pad size of 12. The mean accuracy of Algorithm~\ref{alg:code} and Algorithm~\ref{alg:code_rrc} are 79.64$\pm$0.74 and 75.25$\pm$0.91. There is a drop of 4.39\% of mean accuracy by switching from Algorithm~\ref{alg:code} to Algorithm~\ref{alg:code_rrc}, showing the ineffectiveness of Random-resized-crop on patch classification in the Digital Pathology domain.

\begin{algorithm}[htb]
\caption{Data Augmentation, PyTorch-like}
\label{alg:code_rrc}
\definecolor{codeblue}{rgb}{0.25,0.5,0.5}
\definecolor{codekw}{rgb}{0.85, 0.18, 0.50}
\lstset{
  backgroundcolor=\color{white},
  basicstyle=\fontsize{7.5pt}{7.5pt}\ttfamily\selectfont,
  columns=fullflexible,
  breaklines=true,
  captionpos=b,
  commentstyle=\fontsize{7.5pt}{7.5pt}\color{codeblue},
  keywordstyle=\fontsize{7.5pt}{7.5pt}\color{codekw},
}
\begin{lstlisting}[language=python]
# h,w: height, width; p: padding size
import torchvision.transforms as t

transform = {'train':t.Compose([
                        t.RandomResizedCrop((h, w)),
                        t.RandomHorizontalFlip(0.5),
                        t.ToTensor(),
                        ]),
                 'test':t.Compose([
                        t.Resize((h+p, w+p)),
                        t.CenterCrop((h, w)),
                        t.ToTensor(),
                        ])}

\end{lstlisting}
\end{algorithm}

%\subsection{Experiments without Data Augmentation}

%We also perform experiments without data augmentation (Algorithm~\ref{alg:code_no}) on PCam-Scarce and CRC-Scarce (Figures~\ref{fig:lcrc_1p} and ~\ref{fig:pcam_1p}). 

\begin{algorithm}[htb]
\caption{Without Data Augmentation, PyTorch-like}
\label{alg:code_no}
\definecolor{codeblue}{rgb}{0.25,0.5,0.5}
\definecolor{codekw}{rgb}{0.85, 0.18, 0.50}
\lstset{
  backgroundcolor=\color{white},
  basicstyle=\fontsize{7.5pt}{7.5pt}\ttfamily\selectfont,
  columns=fullflexible,
  breaklines=true,
  captionpos=b,
  commentstyle=\fontsize{7.5pt}{7.5pt}\color{codeblue},
  keywordstyle=\fontsize{7.5pt}{7.5pt}\color{codekw},
}
\begin{lstlisting}[language=python]
# h,w: height, width;
import torchvision.transforms as t

transform = {'train':t.Compose([
                        t.Resize((h, w)),
                        t.ToTensor(),
                        ]),
                 'test':t.Compose([
                        t.Resize((h, w)),
                        t.ToTensor(),
                        ])}

\end{lstlisting}
\end{algorithm}

\section{Conclusion}

%Manipulating the image and input resolution is an effective means to regularize DP models in an annotation-scarce environment.
Input and image resolution are important meta-parameters that have been overlooked to-date in DP classification research. In this paper, we show that tuning the image and input resolution can yield impressive gains in performance. Across annotation-scarce and annotation-rich environments, we demonstrate a positive correlation between the image and input resolution and the patch classification accuracy. By increasing the image and input resolution, our models can capture finer-grained information without losing coarse-grained information, yielding a gain of 6.33\% (PCam-Scarce) and 9.83\% (CRC-Scarce) in patch classification accuracy. %when compared with models trained with the original image and input resolution.
%By exploiting the image and input resolution, we show a gain of 6.33\% and 9.83\% in patch classification accuracy when compared with models trained with the original image and input resolution.
We highlighted other important practical considerations such as the interaction of the data augmentation strategy with resolution and choice of dataset.

\bibliographystyle{IEEEtran}
\bibliography{crv}

\begin{thebibliography}{10}
\providecommand{\url}[1]{#1}
\csname url@samestyle\endcsname
\providecommand{\newblock}{\relax}
\providecommand{\bibinfo}[2]{#2}
\providecommand{\BIBentrySTDinterwordspacing}{\spaceskip=0pt\relax}
\providecommand{\BIBentryALTinterwordstretchfactor}{4}
\providecommand{\BIBentryALTinterwordspacing}{\spaceskip=\fontdimen2\font plus
\BIBentryALTinterwordstretchfactor\fontdimen3\font minus
  \fontdimen4\font\relax}
\providecommand{\BIBforeignlanguage}[2]{{%
\expandafter\ifx\csname l@#1\endcsname\relax
\typeout{** WARNING: IEEEtran.bst: No hyphenation pattern has been}%
\typeout{** loaded for the language `#1'. Using the pattern for}%
\typeout{** the default language instead.}%
\else
\language=\csname l@#1\endcsname
\fi
#2}}
\providecommand{\BIBdecl}{\relax}
\BIBdecl

\bibitem{pantanowitz2010digital}
L.~Pantanowitz, ``Digital images and the future of digital pathology,''
  \emph{Journal of pathology informatics}, vol.~1, 2010.

\bibitem{veeling2018rotation}
B.~S. Veeling, J.~Linmans, J.~Winkens, T.~Cohen, and M.~Welling, ``Rotation
  equivariant cnns for digital pathology,'' in \emph{International Conference
  on Medical image computing and computer-assisted intervention}.\hskip 1em
  plus 0.5em minus 0.4em\relax Springer, 2018, pp. 210--218.

\bibitem{kather2019predicting}
J.~N. Kather, J.~Krisam, P.~Charoentong, T.~Luedde, E.~Herpel, C.-A. Weis,
  T.~Gaiser, A.~Marx, N.~A. Valous, D.~Ferber \emph{et~al.}, ``Predicting
  survival from colorectal cancer histology slides using deep learning: A
  retrospective multicenter study,'' \emph{PLoS medicine}, vol.~16, no.~1, p.
  e1002730, 2019.

\bibitem{aresta2019bach}
G.~Aresta, T.~Ara{\'u}jo, S.~Kwok, S.~S. Chennamsetty, M.~Safwan, V.~Alex,
  B.~Marami, M.~Prastawa, M.~Chan, M.~Donovan \emph{et~al.}, ``Bach: Grand
  challenge on breast cancer histology images,'' \emph{Medical image analysis},
  vol.~56, pp. 122--139, 2019.

\bibitem{srinidhi2020deep}
C.~L. Srinidhi, O.~Ciga, and A.~L. Martel, ``Deep neural network models for
  computational histopathology: A survey,'' \emph{Medical Image Analysis}, p.
  101813, 2020.

\bibitem{krizhevsky2012imagenet}
A.~Krizhevsky, I.~Sutskever, and G.~E. Hinton, ``Imagenet classification with
  deep convolutional neural networks,'' \emph{Advances in neural information
  processing systems}, vol.~25, pp. 1097--1105, 2012.

\bibitem{he2016deep}
K.~He, X.~Zhang, S.~Ren, and J.~Sun, ``Deep residual learning for image
  recognition,'' in \emph{Proceedings of the IEEE conference on computer vision
  and pattern recognition}, 2016, pp. 770--778.

\bibitem{dosovitskiy2020image}
A.~Dosovitskiy, L.~Beyer, A.~Kolesnikov, D.~Weissenborn, X.~Zhai,
  T.~Unterthiner, M.~Dehghani, M.~Minderer, G.~Heigold, S.~Gelly \emph{et~al.},
  ``An image is worth 16x16 words: Transformers for image recognition at
  scale,'' \emph{arXiv preprint arXiv:2010.11929}, 2020.

\bibitem{goodfellow2016deep}
I.~Goodfellow, Y.~Bengio, and A.~Courville, \emph{Deep learning}.\hskip 1em
  plus 0.5em minus 0.4em\relax MIT press Cambridge, 2016, vol.~1.

\bibitem{huh2016makes}
M.~Huh, P.~Agrawal, and A.~A. Efros, ``What makes imagenet good for transfer
  learning?'' \emph{arXiv preprint arXiv:1608.08614}, 2016.

\bibitem{hegde2019similar}
N.~Hegde, J.~D. Hipp, Y.~Liu, M.~Emmert-Buck, E.~Reif, D.~Smilkov, M.~Terry,
  C.~J. Cai, M.~B. Amin, C.~H. Mermel \emph{et~al.}, ``Similar image search for
  histopathology: Smily,'' \emph{NPJ digital medicine}, vol.~2, no.~1, pp.
  1--9, 2019.

\bibitem{teh2020learning}
E.~W. Teh and G.~W. Taylor, ``Learning with less data via weakly labeled patch
  classification in digital pathology,'' in \emph{2020 IEEE 17th International
  Symposium on Biomedical Imaging (ISBI)}.\hskip 1em plus 0.5em minus
  0.4em\relax IEEE, 2020, pp. 471--475.

\bibitem{kupferschmidt2021strength}
K.~L. Kupferschmidt, E.~W. Teh, and G.~W. Taylor, ``Strength in diversity:
  Understanding the impacts of diverse training sets in self-supervised
  pre-training for histology images,'' 2021.

\bibitem{touvron2019fixing}
H.~Touvron, A.~Vedaldi, M.~Douze, and H.~J{\'e}gou, ``Fixing the train-test
  resolution discrepancy,'' \emph{arXiv preprint arXiv:1906.06423}, 2019.

\bibitem{tan2019efficient}
M.~Tan and Q.~Le, ``{E}fficient{N}et: Rethinking model scaling for
  convolutional neural networks,'' in \emph{Proceedings of the 36th
  International Conference on Machine Learning}, ser. Proceedings of Machine
  Learning Research, K.~Chaudhuri and R.~Salakhutdinov, Eds., vol.~97.\hskip
  1em plus 0.5em minus 0.4em\relax PMLR, 09--15 Jun 2019, pp. 6105--6114.

\bibitem{rrc}
``Random-resized crop algorithm - pytorch (source code),''
  \url{https://github.com/pytorch/vision/blob/main/torchvision/transforms/transforms.py#L847},
  accessed: 2022-02-14.

\bibitem{teh2020proxynca++}
E.~W. Teh, T.~DeVries, and G.~W. Taylor, ``Proxynca++: Revisiting and
  revitalizing proxy neighborhood component analysis,'' in \emph{Computer
  Vision--ECCV 2020: 16th European Conference, Glasgow, UK, August 23--28,
  2020, Proceedings, Part XXIV 16}.\hskip 1em plus 0.5em minus 0.4em\relax
  Springer, 2020, pp. 448--464.

\bibitem{xie2017aggregated}
S.~Xie, R.~Girshick, P.~Doll{\'a}r, Z.~Tu, and K.~He, ``Aggregated residual
  transformations for deep neural networks,'' in \emph{Proceedings of the IEEE
  conference on computer vision and pattern recognition}, 2017, pp. 1492--1500.

\end{thebibliography}

\end{document}